\documentclass[lettersize,journal]{IEEEtran}
\usepackage{amsmath,amsfonts}
\usepackage{algorithmic}
\usepackage{algorithm}
\usepackage{array}
\usepackage[caption=false,font=normalsize,labelfont=sf,textfont=sf]{subfig}
\usepackage{textcomp}
\usepackage{stfloats}
\usepackage{url}
\usepackage{verbatim}
\usepackage{graphicx}
\usepackage{cite}
\usepackage{amssymb}
\usepackage{booktabs}
\usepackage{bbm}
\usepackage{multirow}
\usepackage{tabu}
\usepackage{comment}
\usepackage{color}
\newcommand{\ljdcomment}[1]{\textcolor[rgb]{0,0,0} {#1}}
\newcommand{\tabincell}[2]{\begin{tabular}{@{}#1@{}}#2\end{tabular}}

\begin{document}

\title{Layout-Bridging Text-to-Image Synthesis}

\author{Jiadong~Liang, Wenjie~Pei
        and~Feng~Lu,~\IEEEmembership{Member,~IEEE}
	\IEEEcompsocitemizethanks{\IEEEcompsocthanksitem Jiadong Liang and Feng Lu are with the State Key Laboratory of Virtual Reality Technology and Systems, School of Computer Science and Engineering, Beihang University, Beijing 100191, China. Email: $\{$ljdtc, lufeng$\}$@buaa.edu.cn\protect
		\IEEEcompsocthanksitem Wenjie Pei is with the Department of Computer Science, Harbin Institute of Technology at Shenzhen, Shenzhen 518057, China (e-mail:wenjiecoder@outlook.com). \protect
		\IEEEcompsocthanksitem Feng Lu is the corresponding author. E-mail: lufeng@buaa.edu.cn. \protect}
}

\markboth{Journal of \LaTeX\ Class Files,~Vol.~14, No.~8, August~2021}%
{Shell \MakeLowercase{\textit{et al.}}: A Sample Article Using IEEEtran.cls for IEEE Journals}


\maketitle

\begin{abstract}
The crux of text-to-image synthesis stems from the difficulty of preserving the cross-modality semantic consistency between the input text and the synthesized image. 
Typical methods, which seek to model the text-to-image mapping directly, could only capture keywords in the text that indicates common objects or actions but fail to learn their spatial distribution patterns.
An effective way to circumvent this limitation is to generate an image layout as guidance, which is attempted by a few methods. 
Nevertheless, these methods fail to generate practically effective layouts due to the diversity of input text and object location.
In this paper we push for effective modeling in both text-to-layout generation and layout-to-image synthesis. 
Specifically, we formulate the text-to-layout generation as a sequence-to-sequence modeling task, and build our model upon Transformer to learn the spatial relationships between objects by modeling the sequential dependencies between them. 
In the stage of layout-to-image synthesis, we focus on learning the textual-visual semantic alignment per object in the layout to precisely incorporate the input text into the layout-to-image synthesizing process. 
To evaluate the quality of generated layout, we design a new metric specifically, dubbed Layout Quality Score, which considers both the absolute distribution errors of bounding boxes in the layout and the mutual spatial relationships between them. 
Extensive experiments on three datasets demonstrate the superior performance of our method over state-of-the-art methods on both predicting the layout and synthesizing the image from the given text.
\end{abstract}


\section{Introduction}
\IEEEPARstart{T}{ext-to-image} synthesis aims to synthesize a realistic image that is consistent with the textual description. 
It has extensive applications ranging from artistic creation to computer-aided design. Text-to-image synthesis is quite challenging  in that it demands not only 
high quality of the synthesized image, but also cross-modality semantic consistency between the given text and the synthesized image.

\begin{figure}[t]
\centering
    \includegraphics[width=1.0\linewidth]{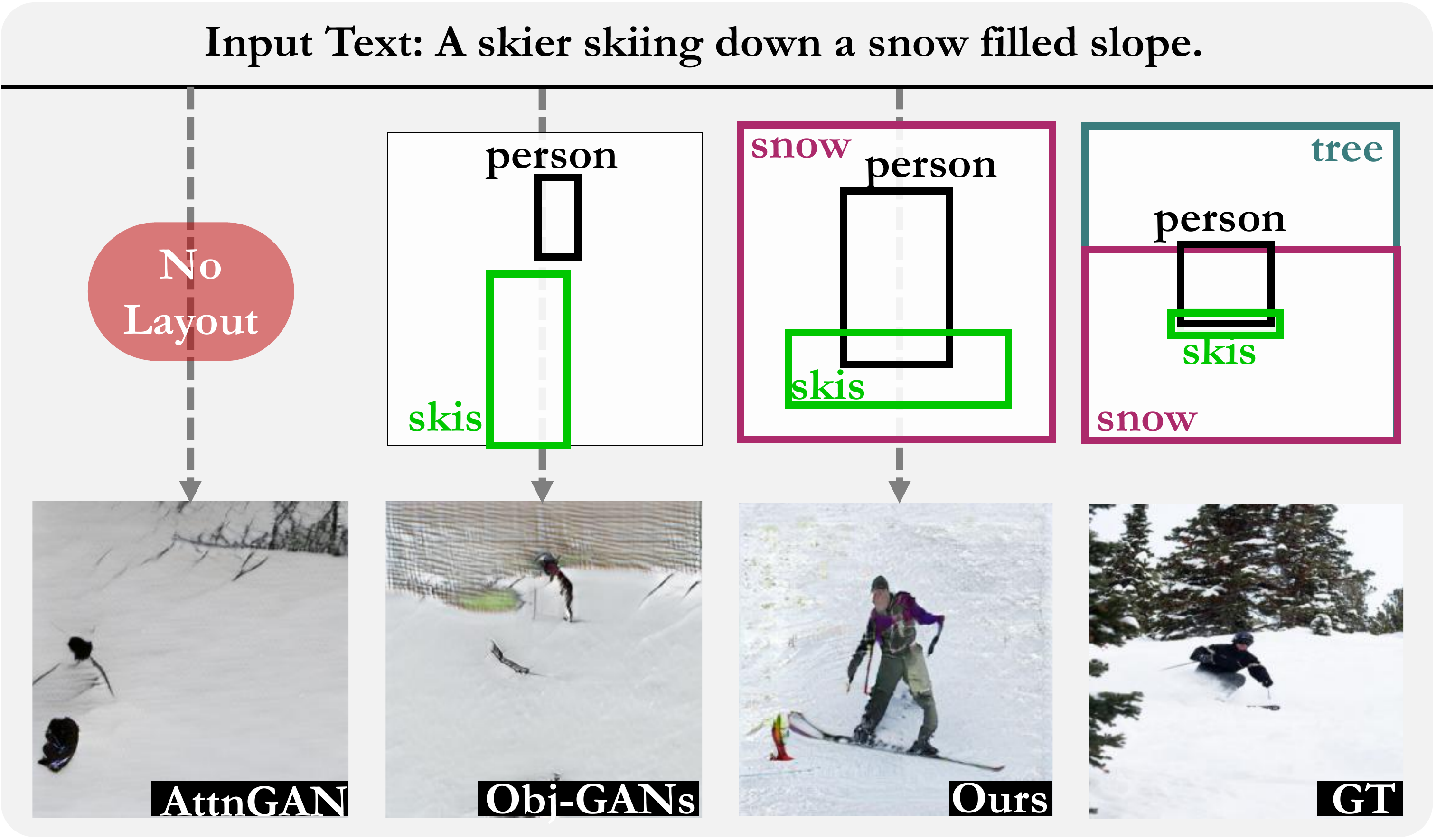}
    \caption{Given the input text, AttnGAN~\cite{xu2018attngan} seeks to synthesize the output image directly while our method and Obj-GANs~\cite{li2019object} generate a layout to bridge the text and the output image. Our method is able to generate more reasonable object layout between 'person' and 'skis' than Obj-GANs~\cite{li2019object}. Guided by the generated layout, our method can synthesize an image with more consistent semantics with the given text than both other methods. } 
\label{Fig:intro}
\end{figure}

Most existing methods~\cite{zhang2017stackgan,xu2018attngan,zhu2019dm,liang2020cpgan,tao2020df, ruan2021dae} for text-to-image synthesis focus on modeling direct text-to-image semantic mapping by incorporating the input textual information into the generative processing. 
Despite the substantial progress made by these methods, an important limitation is that such methods can only capture keywords in the text indicating typical objects or actions, but fail to learn the spatial distribution patterns of objects in the image, namely the image layout. This is basically resulted from the difficulties of modeling the text-to-image mapping directly. As a result, the synthesized images do not have a reasonable object layout with correct spatial relationships between objects. As an example, AttnGAN~\cite{xu2018attngan} in Figure~\ref{Fig:intro} could only synthesize an image containing texture of person but fail to capture a reasonable layout between `person' and `skis'.

\noindent An intuitive and straightforward way to tackle above limitation is to generate an image layout that is consistent with the given text, and then use the layout as a transitional modality to bridge the gap between the input text and the synthesized image. A state-of-the-art method following this way is Obj-GANs~\cite{li2019object}, which employs a bi-directional LSTM~\cite{huang2015bidirectional} to learn hidden representations for the input text and then generates a layout by training another LSTM. Whilst Obj-GANs can obtain plausible layout in some cases, the generated layout tends to over-fit simple patterns, which cannot reflect the diversity in real scenarios due to its limited capability of modeling object layout. 
Besides, it is also arduous for Obj-GANs to learn the complex spatial relationships between objects. As shown in Figure~\ref{Fig:intro}, the generated layout by Obj-GANs fails to indicate the correct spatial relationship between `person' and `skis'. 

In this paper, we decompose the text-to-image synthesis task into two easier sub-tasks: text-to-layout generation and layout-to-image synthesis, and push for the effective modeling in both two sub-tasks. To learn the text-to-layout mapping precisely, we formulate the layout generating process as a sequential prediction of objects and model the objects' relationships as sequential dependencies. As a result, the text-to-layout generation boils down to a sequence-to-sequence modeling task. We adopt Transformer~\cite{vaswani2017attention} as the internal encoding and decoding structure due to its powerful capability of sequence-to-sequence modeling. The pivotal point to adapt Transformer to text-to-layout generation lies in how to predict the category and the location for each object in the layout correctly. We perform a unified prediction scheme, in which we partition the layout area into $S\times S$ grids equally and predict the object category and its center position jointly.
Such unified scheme is not only efficient for prediction, but also enables our model to better perceive the object state and model the object dependencies by associating the category and location together. 

The generated layout is further leveraged to guide the image synthesis in the stage of layout-to-image synthesis. To be specific, we draw on the generative structure of LostGANs\_V2~\cite{sun2020learning} and focus on learning the textual-visual semantic alignment per object in the layout to precisely incorporate the input textual description into the layout-to-image synthesizing process. 

The presented example in Figure~\ref{Fig:intro} shows that our model is able to generate much more reasonable layout than ObjGANs, and thereby synthesize more consistent image with the given textual description than other two methods. To conclude, we make following contributions.
\begin{itemize}
\item We design a specific model based on Transformer to generate high-quality image layout which is consistent with the input text. 
\item A layout-to-image synthesizer is proposed to  synthesize the image by incorporating both the generated layout and the input text. 
\item We propose quantitative metrics for evaluating the quality of generated layout. 
\item Extensive experiments on three datasets demonstrate that our proposed method performs favorably against state-of-the-art methods on both predicting the image layout and synthesizing the image from the given text. 
\end{itemize}

\section{Related Works}
\subsection{Text-to-image synthesis without layout.}
Generative adversarial network (GAN)~\cite{goodfellow2014generative} has been widely employed for text-to-image synthesis. Representative methods include: 
Reed~et al.~\cite{reed2016generative} that proposes an early work for text-to-image synthesis at low resolutions,
StackGAN~\cite{zhang2017stackgan} which achieves high resolution image synthesis with a coarse-to-fine framework,
AttnGAN~\cite{xu2018attngan} leveraging word-level features as input to perform fine-grained image generation by employing attention mechanism,
DMGAN~\cite{zhu2019dm} that introduces a dynamic memory module to fix distortions of initial images that usually cause large errors in the subsequent generations, and 
MirrorGAN~\cite{qiao2019mirrorgan} which first adopts a text-to-image-to-text cycle framework to guarantee text-image consistency. Based on these fundamental works, the research on text-to-image synthesis is further advanced on different aspects, as discussed below.

\noindent\textbf{Parsing the semantics of input text.} Some works~\cite{lao2019dual,cheng2020rifegan,qiao2019learn,ruan2021dae,park2021benchmark} further optimized the details of the generated images by parsing the input text.
Lao~et al.~\cite{lao2019dual} disentangle the input text into two types latent code of content and style by a dual adversarial inference mechanism to obtain comprehensive semantic information.
SD-GAN~\cite{yin2019semantics} further improved the image synthesis quality by exploring the consistency between the input sentences.
RiFeGAN~\cite{cheng2020rifegan} used multiple sentences as input to provide rich semantic information to the generator.
LeicaGAN~\cite{qiao2019learn} proposed text-visual co-embeddings to translate input text to corresponding visual features. 
DAE-GAN~\cite{ruan2021dae} adopts aspect information of text to enhance the details of synthesized images.
Park~et al.~\cite{park2021benchmark} proposed new benchmarks to perform a systematic study of the model's generalization performance for novel word compositions.

\noindent\textbf{Modeling the generators and discriminators.} Many studies~\cite{liu2020time,tan2019semantics,cha2019adversarial,li2019controllable,yin2019semantics} have focused on developing delicate discriminators to generate images that are well-matched to the input text.
SEGAN~\cite{tan2019semantics} combined conditional discriminators with the Siamese networks to achieve fine-grained text-to-image generation.
Text-SeGAN~\cite{cha2019adversarial} can generate diverse images that are semantically relevant to the input text by adding a semantic classifier to the discriminator.
ControlGAN~\cite{li2019controllable} introduced fine-grained word semantic as conditional information in the discriminator by an attention mechanism to push for the text-image semantic alignment.
TIME~\cite{liu2020time} adopts a image caption model as discriminator to improve semantic consistency.

However, to generate high quality images, these methods generally employed stacked generative adversarial networks, which further increases the computational consumption and reduces training efficiency.
DT-GAN~\cite{9533527} and DF-GAN~\cite{tao2020df} only employed a pair of generator and discriminator to synthesize high-quality and semantically consistent images.
CKD~\cite{yuan2019ckd} designed a hierarchical knowledge distillation paradigm to extract image semantic which injects into the image generation process.

Another interesting line of research has explored the generation of faces and food.
CookGAN~\cite{zhu2020cookgan} completed the step-by-step generation of food in an interactive way.
ChefGAN~\cite{pan2020chefgan} performed recipe-to-food generation by a joint image-recipe embedding model.
SEA-T2F~\cite{sun2021multi} is proposed for text-to-face generation, where multiple captions are used as inputs to improve the semantic consistency of the generated faces.
TediGAN~\cite{xia2021tedigan} proposed a novel framework for multi-modal face generation and manipulation with textual descriptions.
Zhou~et al.~\cite{9666791} employed a pre-trained transformer-based BERT model and a StyleGAN encoder to synthesize the realistic face images.

\noindent\textbf{Generating diverse objects.} Although these aforementioned approaches have achieved significant progress on some easy datasets which only have one object per image (e.g. bird, flower, face, and food).
The context-rich image synthesis which has complex objects is still challenging.
CPGAN~\cite{liang2020cpgan} focused on parsing the content of both the input text and the synthesized image thoroughly to model the text-to-image consistency of complex image.
Huang~\cite{huang2021unifying} designed a unified image-and-text framework to jointly study image-to-text and text-to-image generations.
XMC-GAN~\cite{zhang2021cross} efficiently performs cross-modal translation by maximizing the mutual information between image and text.
Since the semantic ambiguity of the input text and the image diversity, it is difficult for these methods to generate images with correct spatial relationships of objects.

\noindent\textbf{Leveraging Transformer instead of CNNs as the backbone.} In recent years, thanks to the rapid development of VIT~\cite{dosovitskiy2020image}, several works have attempted to solve the text-to-image generation problem by training directly on large-scale datasets in an end-to-end manner, such as DALL-E~\cite{ramesh2021zero} and CogView~\cite{ding2021cogview}.
The latest works of Imagen~\cite{imagen}, DALL-E2~\cite{DALLE2} and GLIDE~\cite{nichol2021glide} also demonstrate the potential of diffusion models for text-to-image synthesis.
To achieve good visualization results, these methods not only have a huge number of trainable parameters but also require a gigantic scale of labeled data for training.
In other words, they typically require hundreds of GPUs and weeks of training time to get the desired results, which limits further development in this field.

\begin{figure*}[t]
\centering
\includegraphics[height=6.4cm]{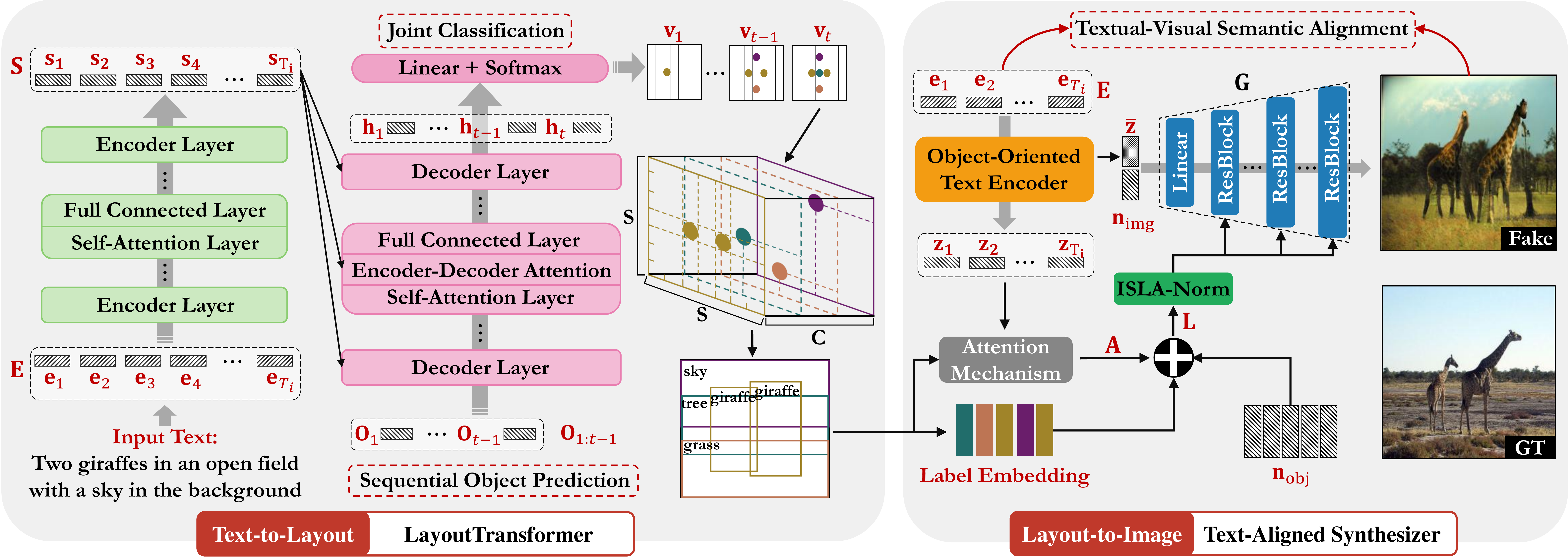}
\caption{Architecture of the proposed method, which performs text-to-image synthesis in two stages: text-to-layout Generation by the proposed LayoutTransformer and layout-to-image synthesis by the proposed Text-Aligned Layout-to-Image Synthesizer.}
\label{fig:model}
\end{figure*}

\subsection{Text-to-image synthesis with image layout.}
Image layout captures the spatial distribution patterns of objects in the scene and thus has been taken into consideration by a few recent text-to-image works.
Hong~et al.~\cite{hong2018inferring} first developed a text-to-layout-to-image framework to generate images with reasonable spatial distributions.
Following this work, Obj-GANs~\cite{li2019object} further improved the layout synthesis quality by incorporating fine-grained word embeddings.
By employing the layout synthesis from Obj-GAN~\cite{li2019object}, OP-GAN~\cite{hinz2019semantic} proposed a new spatial location related generator to achieve text-to-image synthesis.
R-GAN~\cite{qiao2021r} adopts a similar method of text-to-layout generation as Obj-GAN and further incorporates scene graph information to synthesize more reasonable layouts.
Besides, Text2Scene~\cite{tan2019text2scene} focused on text-to-layout synthesis by designing a Seq-to-Seq framework.
Among these recent works, Obj-GAN is a representative that performs well in many simple scenarios. However, modeling complex spatial relationship between objects still remains challenging.

\section{Method}
Given a textual description, we aim to generate a high-quality layout that indicates the spatial distribution of objects in the image to be synthesized. 
The generated layout, which is expected to bridge the gap between the input text and the synthesized image, is further leveraged to guide the image synthesis and thereby yielding high-quality image for the given text. Thus, our method performs text-to-image synthesis in two stages: text-to-layout generation and layout-to-image synthesis.
Figure~\ref{fig:model} illustrates the overall structure of our method.
\subsection{Text-to-Layout Generation}
The layout is generated following an encoder-decoder framework by our proposed LayoutTransformer, which consists of two modules: Text Encoder and Layout Decoder. It first encodes the input text to learn latent representations for the entire text by its Text Encoder, then employs Layout Decoder to generate the image layout. We formulate the layout generating process as a sequential prediction of objects. As a result, the layout generation boils down to a sequence-to-sequence modeling task. Hence we design the text-to-layout generation by adopting the similar internal structure as Transformer~\cite{vaswani2017attention} due to its powerful capability of sequence-to-sequence modeling.

\subsubsection{Text Encoder}
\label{Sec:text_encoder}
The Text Encoder of LayoutTransformer performs text encoding in the similar way as the encoder of classical Transformer, which takes a sequence of word embeddings as input and learns a corresponding sequence of representations with the same length as the input sequence:
\begin{equation}
    \{\mathbf{s}_1, \dots, \mathbf{s}_{T_i}\} = \mathcal{F}_{e}(\{\mathbf{e}_1, \dots, \mathbf{e}_{T_i}\}),
\end{equation}
where $T_i$ is the length (in word) of the input text. $\mathcal{F}_e$ denotes the transformation function performed by Text Encoder, which is composed of 6 identical basic layers. Each basic layer comprises a multi-head self-attention layer and a position-wise fully connected layer. The input word embedding at $t$-th time step $\mathbf{e}_t \in \mathbb{R}^D$ corresponds to the indexed vectorial representation in the embedding matrix $\mathbf{E} \in \mathbb{R}^{D \times K}$ for a word dictionary containing $K$ words, which is learned together with the whole model.

\begin{figure}[ht]
\centering
\includegraphics[height=2.4cm]{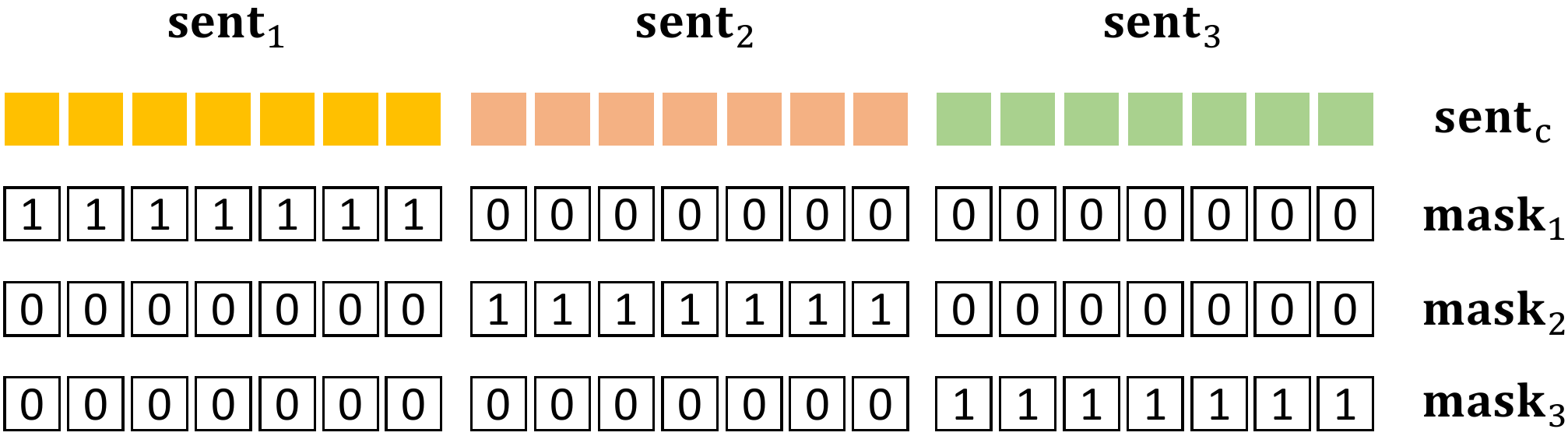}
\caption{Visualization of synthesized multi-caption masking.}
\label{fig:multi_caption}
\end{figure}

\noindent{\textbf{Incorporation of multi-caption input}.}
The captions for a same image are highly diverse, it means that naively using these captions as unique inputs will generate unstable layouts.
Hence, we concatenate them together as input to form a more comprehensive description.
To avoid semantic inter-pollution between different initial captions, 
we perform masking to ensure that each word in the concatenated caption can only attend to words in the same initial caption while other words are masked out to be invisible as shown in Figure~\ref{fig:multi_caption}.
Specifically, given three initial input captions$\{\mathbf{sent}_1, \mathbf{sent}_{2}, \mathbf{sent}_{3}\}$, we construct corresponding masks $\{\mathbf{mask}_1, \mathbf{mask}_{2}, \mathbf{mask}_{3}\}$ for each of initial input captions.
When we encoder the word in $\mathbf{sent}_i$, we perform element-wise multiplication between the concatenated input text $\mathbf{sent}_{c}$ and the corresponding mask $\mathbf{mask}_i$ to ensure that each word in the concatenated caption can only attend to words in the same initial caption while other words are masked out to be invisible.

\subsubsection{Layout Decoder}
The obtained latent representations $\mathbf{S} = \{\mathbf{s}_1, \dots, \mathbf{s}_T\}$ from Text Encoder are then decoded by Layout Decoder of our LayoutTransformer to generate the image layout. 
The key difference between our method and Transformer is that the output of our Layout Decoder is image layout instead of a text sequence. Thus, how to formulate the layout generating process is crucial to the design of Layout Decoder.

\noindent\textbf{Sequential object prediction.} The generated layout is required to show the object distribution in the image to be synthesized. Thus generating an image layout involves predicting the bounding box (including both location and size) and the class label for each potential object. Besides, the dependencies between objects, such as spatial or semantic associations, should be taken into account during the layout generation. Hence the prediction of current object should depend on the previously predicted objects rather than predicting objects independently.
Based on these two considerations, we design Layout Decoder to predict the objects sequentially to generate the image layout progressively, i.e, predicting one object per step. Accordingly, we adopt similar sequential decoding structure as Transformer for Layout Decoder, which also employs 6 basic layers consisting of multi-head attentions and fully connected layers for decoding. Formally, Layout Decoder performs decoding to learn the hidden state $\mathbf{h}_t$ at $t$-th time step prepared for output by:
\begin{equation}
\mathbf{h}_t = \mathcal{F}_d(\mathbf{S}, \mathbf{O}_{1:{t-1}}). 
\label{eqn:decode}
\end{equation}
Herein, $\mathbf{O}_{1:{t-1}}$ denotes the previously predicted results and $\mathcal{F}_d$ refers to the transformation function by the decoding process of Layout Decoder.

\noindent\textbf{Joint classification of object category and position.} 
For prediction of the each object in the layout, Layout Decoder needs to predict not only the class label of the object, but the size and location of the bounding box.
We find that although the distribution of objects in a layout is highly diverse, the location of objects is directly related to their category.
Hence, predicting the class label of objects and the bounding boxes jointly can facilitate the predicting precision.
Specifically, we partition the layout area into a $S \times S$ grid, and then predict the class label and the grid cell containing the object center jointly each time we predict an object. Consequently, we formulate it as a  classification problem with total $S\times S \times C$ classes, where $C$ is total category number of objects. It is equivalent to evaluating all possibilities of combination of the category and the location (in grid) for the object to be predicted and then finding the optimal option, namely the grid cell with highest confidence that contains the most probable category of object.
Formally, Layout Decoder performs classification at $t$-th time step based on the hidden states $\mathbf{h}_t$ by:
\begin{equation}
\begin{split}
    &v_t = \underset{i \in [1, S \times S \times C]}{\text{argmax}(\mathbf{p}_t^i)}, \quad \mathbf{p}_t = \text{Softmax}(\mathbf{M} \mathbf{h}_t),
    \label{eqn:cls}
\end{split}
\end{equation}
where $\mathbf{M}$ denotes a linear transformation. $\mathbf{p}_t \in \mathbb{R}^{S \times S \times C}$ is calculated probabilities for total $S \times S \times C$ classification entries. $v_t$ is the entry with the maximal probability, from which we can infer the corresponding class label $c_t$ of the predicted object in this step as well as the grid cell $<G_t^x\in [1, S], G_t^y \in [1, S]>$ containing the object center. Note that $\mathbf{O}_t$ in Equation~\ref{eqn:decode} is represented in one-hot encoding of $v_t$.

\ljdcomment{
Finally, we directly regress vertex coordinates of the bounding box based on the grid cell $<G_t^x, G_t^y>$.
The classification performed in Equation~\ref{eqn:cls} predicts the grid cell that the object center fall in, which is a coarse object position. To predict the precise location of bounding box center within the localized grid cell as well as the size of the bounding box, we performs fine-grained regression by a regression head (denoted as $\mathcal{F}_{\text{reg}}$) consisting of three fully connected layers as well as activation function ReLU to predict the state of bounding box $\mathbf{f}_t = \{f_t^x, f_t^y, f_t^w, f_t^h\}$ at $t$-th step:
\vspace{-3pt}
\begin{equation}
\vspace{-3pt}
\{f_t^x, f_t^y, f_t^w, f_t^h\}= \mathcal{F}_{\text{reg}}(\mathbf{h}_t),
\label{eqn:reg}
\end{equation}
where $f_t^x, f_t^y$ are coordinates of bounding box center within the grid cell $<G_t^x, G_t^y>$, and $f_t^w, f_t^h$ jointly indicate the bounding box size of the predicted object.
Such coarse-to-precise localization strategy decomposes the complex localization problem into two much easier prediction steps.}

A prominent benefit of formulating the prediction of both the category and position of the object as one integrated classification task is that such mechanism enables the model to achieve more accurate estimation of the object state than predicting the category and position separately. 
As a result, such mechanism facilitates the modeling of dependencies between predicted objects in sequential object prediction shown in Equation~\ref{eqn:decode}, which is validated in experiments in Section~\ref{Sec:ablation}.

\subsubsection{Parameter Learning}
\label{Sec:layout_training}
The whole model of LayoutTransformer for text-to-layout generation, including Text Encoder and Layout Decoder, is trained under the supervision of two types of losses: the classification loss $\mathcal{L}_{\text{cls}}$ for prediction in Equation~\ref{eqn:cls} and regression loss $\mathcal{L}_{\text{reg}}$ for bounding boxes regression: 

\begin{equation}
\begin{split}
    \mathcal{L}_{\text{layout}} = \mathcal{L}_{\text{cls}} + \lambda \mathcal{L}_{\text{reg}},
    \mathcal{L}_{\text{cls}} = \sum_{n=1}^N\sum_{t=1}^{T_o^{(n)}}\mathcal{L}_{\text{CE}}(\mathbf{p}_t, \hat{\mathbf{p}_t}),
\label{eqn:reg}
\end{split}
\end{equation}
where $\mathcal{L}_{\text{CE}}$ is cross entropy loss and $\mathcal{L}_{\text{reg}}$ is the regression loss of bounding box adopted in YOLO-V3~\cite{redmon2018yolov3}. Here ${T_o^{(n)}}$ denotes the predicted sequence length, namely the number of objects, for the $n$-th training sample composed of a textual description and a paired image. $\lambda$ is the hyper-parameter to balance two losses.
\subsection{Layout-to-Image Synthesis}
The generated layout is further leveraged to guide the image synthesis, thereby synthesizing an image that is consistent with both the generated layout and the textual description.
As shown in Figure~\ref{fig:model}, we draw on generative structure of LostGAN\_V2~\cite{sun2020learning} and build our image synthesizer upon it. Since the goal of LostGAN\_V2 is to synthesize image from sole layout without textual description, we incorporate the textual description into our image synthesizer to ensure the textual-visual consistency. 
\subsubsection{Text-Aligned Layout-to-Image Synthesizer}
Text-Aligned Layout-to-Image Synthesizer (\emph{TALIS}) is designed to generate the output image, taking both the given textual description and the generated layout as input. As illustrated in Figure~\ref{fig:model}, its basic backbone is the generative network $\mathbf{G}$, which consists of one fully connected layer and five residual building blocks (ResBlocks). The image synthesis is performed by two core operations. 1) \textbf{Basis image generation}: the generative network $\mathbf{G}$ takes an initial vectorial embedding $\mathbf{q}_{\text{img}}$ to generate the basis of the whole image. 2) \textbf{Guidance of layout}: a feature normalization module termed ISLA-Norm of LostGAN\_V2 is employed to steer the generating process of $\mathbf{G}$ based on the generated layout to achieve the layout-image consistency. The semantics of the input text are incorporated into both operations.

Since the Text Encoder designed for text-to-layout generation aims to learn textual embeddings that are particularly used for predicting the layout, it is not well suitable for layout-to-image synthesis in which textual-visual semantic alignment is crucial. Thus, we learn an individual text encoder to extract textual semantics for describing each object in the layout. The proposed text encoder, referred to as Object-Oriented Text Encoder, utilizes the encoder of transformer $\mathcal{E}_T$ to learn the hidden representations for each word in the text:
\begin{equation}
    \{\mathbf{z}_1, \dots, \mathbf{z}_{T_i}\} = \mathcal{E}_T(\{\mathbf{e}_1, \dots, \mathbf{e}_{T_i}\}),
\end{equation}
where word embedding at $t$-th step $\mathbf{e}_t \in \mathbb{R}^D$ is the vectorial representation indexed from a learned embedding matrix $\mathbf{E} \in \mathbb{R}^{D\times K}$. 

\noindent\textbf{Basis image generation.} 
The global representation $\bar{\mathbf{z}}$ for the whole textual description is obtained by averaging word features $\{\mathbf{z}_1, \dots, \mathbf{z}_{T_i}\}$.
It is concatenated with a noise embedding $\mathbf{n}_{\text{img}}$ sampled from a normal distribution to form the initial embeddings  $\mathbf{q}_{\text{img}}$ for $\mathbf{G}$:
\begin{equation}
\mathbf{q}_{\text{img}} = \text{Concat}(\bar{\mathbf{z}}, \mathbf{n}_{\text{img}}).
\end{equation}
\noindent\textbf{Guidance of layout.} The learned textual representations are also incorporated into the feature normalization module ISLA-Norm to manipulate the generating process of $\mathbf{G}$. Specifically, we extract the related semantics for each object in the generated layout by attending to each word in the text and measuring their compatibility to the object category. For instance, the textual semantics $\mathbf{a}_k$ for $k$-th object  in the layout with the class label $c_k$ are extracted by:
\begin{equation}
\resizebox{0.8\linewidth}{!}{$
    \begin{split}
        & \mathbf{a}_k = \sum_{t=1}^{T_i} w_t \mathbf{z}_{t},\quad w_t = \frac{\exp({\mathbf{z}_{t}^\top \mathbf{U}[c_k, :]})}{\sum_{t=1}^{T_i}\exp{\mathbf{z}_{t}^\top \mathbf{U}[c_k, :]}},
    \end{split}
    $}
    \label{eqn:text_semantic}
\end{equation}
where $\mathbf{U} \in \mathbb{R}^{C \times d_c}$ is a learned embedding matrix for all $C$ categories. $w_t$ denotes the compatibility between the object category $c_k$ and the $t$-th word. In this way we can learn the textual semantics for all $T_o$ objects in the layout $\mathbf{A} = \{\mathbf{a}_1, \dots, \mathbf{a}_{T_o}\}$.
The obtained textual semantics are concatenated with the category embeddings and a noise embedding together for each object to form the joint input vector for ISLA-Norm:
\begin{equation}
\begin{split}
    &\mathbf{l}_k = \text{Concat}(\mathbf{a}_k, \mathbf{U}[c_k, :], \mathbf{n}_{\text{obj}}^k),\ k= 1, \dots, T_o,
\end{split}
\end{equation}
where $\mathbf{n}_{\text{obj}}^k$ is a sampled noise embedding for $k$-th object. ISLA-Norm utilizes the obtained layout information $\mathbf{L} = \{\mathbf{l}_1, \dots, \mathbf{l}_{T_o}\}$ to perform feature normalization for feature maps in each ResBlock of the generative network $\mathbf{G}$. The details of ISLA-Norm are presented in LostGAN\_V2~\cite{sun2020learning}. The generative network $\mathbf{G}$ is trained following the training procedures of LostGAN\_V2 while the proposed Object-Oriented Text Encoder is pre-trained individually, which is explicated in the following subsection.

\subsubsection{Learning the Textual-Visual Semantic Alignment}
One key point that affects the performance of the proposed Text-Aligned Layout-to-Image Synthesizer (\emph{TALIS}) is that the learned textual semantics for each object by Object-Oriented Text Encoder should be aligned well with the corresponding visual semantics corresponding to the synthesized image for the same object. To this end, we pre-train Object-Oriented Text Encoder individually to push for the textual-visual semantic alignment. 
Given a text-image pair($X, I$) for training, we extract the visual features for each object in the image by a pre-trained VGG net~\cite{simonyan2014very}.
\begin{equation}
    \mathbf{B} ,\Bar{\mathbf{b}} = \text{ROI}(\text{VGG}(I)),
\label{eqn:ROI}
\end{equation}
where $\mathbf{B} = \{\mathbf{b}_1, \dots, \mathbf{b}_{T_o}\}$ is the visual features for total $T_o$ object and $\Bar{\mathbf{b}}$ is the visual feature for the whole image. Herein ROI is the ROI-align layer appended after the feature maps of VGG net for cropping features for objects. We aim to maximize the consistency between the visual features $\mathbf{B}$ and the textual semantics $\mathbf{A}$ obtained in Equation~\ref{eqn:text_semantic} for all objects in the image. Thus we define the consistency score between them based on Cosine distance:
\begin{equation}
\text{S}_{\text{obj}} (X, I) = \text{log}(\sum_{k=1}^{T_o}\text{exp}(\frac{\mathbf{b}_k \mathbf{a}_k}{\|\mathbf{b}_k\|\|\mathbf{a}_k\|})).
\label{eqn:object-text_score}
\end{equation}
Similarly, we also measure the global consistency score between the global textual semantics $\Bar{\mathbf{e}}$ and the global image features $\Bar{\mathbf{b}}$:
\begin{equation}
\text{S}_{\text{img}}(X,I) = \frac{\Bar{\mathbf{b}} \Bar{\mathbf{e}}}{\|\Bar{\mathbf{b}}\|\|\Bar{\mathbf{e}}\|}).
\label{eqn:image-text_score}
\end{equation}
Finally, we train the Object-Oriented Text Encoder by a contrastive loss,
\begin{equation}
\resizebox{0.65\hsize}{!}{$
\begin{split}
& \mathcal{L}_{\text{con}} = \mathcal{L}_{\text{obj}} + \mathcal{L}_{\text{img}}, \\
& \mathcal{L}_{\text{obj}} = -\text{log}\frac{\text{exp}(\text{S}_{\text{obj}}(X_k, I_k))}{\sum_{i=1}^{J}\text{exp}(\text{S}_{\text{obj}}(X_i, I_k))}, \\
& \mathcal{L}_{\text{img}}= -\text{log}\frac{\text{exp}(\text{S}_{\text{img}}(X_k, I_k))}{\sum_{i=1}^{J}\text{exp}(\text{S}_{\text{img}}(X_i, I_k))}, \\
\end{split}
$}
\end{equation}
where $J$ is the batch size in the training stage.

\section{Experiment}
\subsection{Experimental Setup}
\noindent\textbf{Dataset.}
We use three datasets to evaluate our method. 1) COCO~\cite{lin2014microsoft}, 2) COCO-stuff~\cite{caesar2018coco} and 3) LN-COCO. COCO~\cite{lin2014microsoft} dataset is commonly used for text-to-image synthesis. Each image has 5 corresponding textual captions. Compared with the 80 instance categories (bicycle, bus, etc.) in COCO dataset, COCO-stuff~\cite{caesar2018coco} is a more challenging dataset containing additional 91 stuff categories (hill, grass, etc.). LN-COCO has the same set of images as COCO-stuff whilst each image is described by a narrative obtained by Localized Narratives~\cite{pont2020connecting}. Narratives are four times longer than COCO and COCO-stuff captions on average and thus are more informative, which makes text-to-layout synthesis on LN-COCO much more challenging than on COCO and COCO-stuff.

\noindent\textbf{Implementation details.}
Since the gradient cannot be back-propagated through the generated layout, we train the LayoutTransformer for text-to-layout generation and the \emph{TALIS} for layout-to-image synthesis separately. Both networks are trained from scratch.
In our experiment, the grid size $S$ of the Layout Transformer is set to be 7. 
The hyper-parameter $\lambda$ in Equation~\ref{eqn:reg} is set to be 2.0. 
All networks are trained using Adam~\cite{kingma2014adam}.

\subsection{Evaluation Metrics}
We employ existing standard metrics for the evaluation of synthesized images, and propose a new metric for the evaluation of text-to-layout generation due to the lack of effective metrics.
 \subsubsection{Metrics for Evaluating Synthesized Images.} 
 We adopt the five standard metrics for evaluating the quality of synthesized images: Inception Score~\cite{salimans2016improved} (\textbf{IS}), Frechet Inception Distance~\cite{heusel2017gans} (\textbf{FID}), \textbf{R-precision}~\cite{xu2018attngan}, \textbf{ClipScore}~\cite{huang2021unifying} and \textbf{SOA}~\cite{hinz2019semantic}. IS focuses on evaluating the authenticity and diversity of synthesized images while FID measures the distribution distance between synthesized images and the corresponding groundtruth images. 
 R-precision usually uses a vision-and-language retrieval model to measure the semantic consistency between the textual description and the synthesized image. However, the retrieval models used in previous works are inconsistent, which can lead to biased behavior. In this paper, we uniformly use a powerful vision-and-language
 CLIP~\cite{radford2021learning} model as the retrieval model to accomplish the quantitative evaluation of semantic consistency. Compared to R-precision, which evaluates semantic similarity by ranking retrieval results, ClipScore adopts the CLIP model to directly calculate the cosine similarity of text-image pairs. In this paper, we take Clipscore as a fine-grained semantic consistency evaluation metric to complement R-precision.
 SOA adopts a pre-trained object detection network to measure whether the objects specifically mentioned in the caption are recognizable in the generated images.
 Specifically, it includes two sub-metrics: SOA-C (average recall w.r.t. object category) and SOA-I (average recall w.r.t. image sample).

\subsubsection{Metrics for Evaluating Generated Layouts.}
DocSim~\cite{patil2020read} was proposed to evaluate the quality of generated layouts. However, Docsim only measures the absolute errors of bounding box positions between the generated layout and the ground truth while neglecting the mutual spatial relationships between bounding boxes in a layout. 
Thus, we propose a new metric specifically for evaluating the generated layouts called Layout Quality Score \textbf{(LQS)}, which considers both the absolute distribution errors of bounding boxes and the mutual spatial relationships between them.
Concretely, we consider four measurements as follows.

\noindent\textbf{Label Recall (LR)}, which measures the recall rate of the object categories in the generated layout referring to the groundtruth layout. 

\noindent\textbf{Label Precision (LP)}, measuring the precision of the object categories in the generated layout compared to the groundtruth layout. Formally, We denote the label set of the objects in a generated layout containing $\hat{T}_o$ objects as $\hat{\mathbf{d}} = \{\hat{d}_1, \hat{d}_2, \dots, \hat{d}_{\hat{T}_o}\}$
 Similarly, the label set in the corresponding groundtruth layout is denoted as $\mathbf{d} = \{d_1, d_2, \dots, d_{T_o}\}$. 
\textbf{LR} and \textbf{LP} are defined as:
\begin{equation}
\begin{split}
    \text{LR} = \frac{\mid \mathbf{d}  \cap \hat{\mathbf{d}} \mid}{T_o}, \quad
    \text{LP} = \frac{\mid \mathbf{d}  \cap \hat{\mathbf{d}} \mid}{\hat{T_o}}.
    \end{split}
\end{equation}
\noindent\textbf{Location Consistency (LC)}, which measures the location consistency between the objects in the generated layout and those in the groundtruth layout with the same category. We consider two kinds of consistency: 1) Absolute Location Consistency (ALC) which is defined as the distance between the predicted object centers to the corresponding groundtruth object centers and 2) Relative Location Consistency (RLC) which measures the consistency of pairwise distance between two objects, comparing the generated layout with the groundtruth layout. Denoting $\mathbf{m} = \mathbf{d} \cap \hat{\mathbf{d}}$, they are defined as:
\begin{equation}
\resizebox{0.7\hsize}{!}{$
\begin{split}
    & \text{ALC} = \frac{\sum_{o\in \mathbf{m}} \parallel \mathbf{g}_o - \hat{\mathbf{g}}_o \parallel_2}{\mid \mathbf{m} \mid},\\
    & \text{RLC} = \frac{\sum_{i \in \mathbf{m}}\sum_{j\in \mathbf{m} \ \& \ i \neq j}\parallel \mathbf{r}_{i,j} - \hat{\mathbf{r}}_{i,j} \parallel_2}{\mid \mathbf{m} \mid\times(\mid \mathbf{m} \mid-1)},
    \label{eqn:ACD_RCD}
\end{split}
$}
\end{equation}
where $\mathbf{g}_o$ and $\hat{\mathbf{g}}_o$ are the coordinates of the $o$-th object in intersection set $\mathbf{m}$ (correctly predicted objects). $\mathbf{r}_{i,j} = \mathbf{g}_i - \mathbf{g}_j$ is the relative distance between the $i$-th object and the $j$-th object.
However, the values of ALC and RLC are directly related to the whole image area $u_s$ (the larger $u_s$, the larger the values of ALC and RLC). To remove the effect of $u_s$ on the value of ALC and RLC, we adopt the Gaussian kernel function (parameterized by $\sigma_l$) to calculate smoothing values. Finally, we define Location Consistency (\textbf{LC}) by combining these two measurements:
\begin{equation}
\resizebox{0.88\hsize}{!}{$
    \text{LC} = \gamma_{\text{lc}}\text{exp}(-\text{ALC}/2\sigma_{l}^2) + (1 - \gamma_{\text{lc}})\text{exp}(-\text{RLC}/2\sigma_{l}^2).
    $}
\label{eqn:LDS}
\end{equation}

\begin{figure*}[t]
\centering
\includegraphics[height=3.4cm]{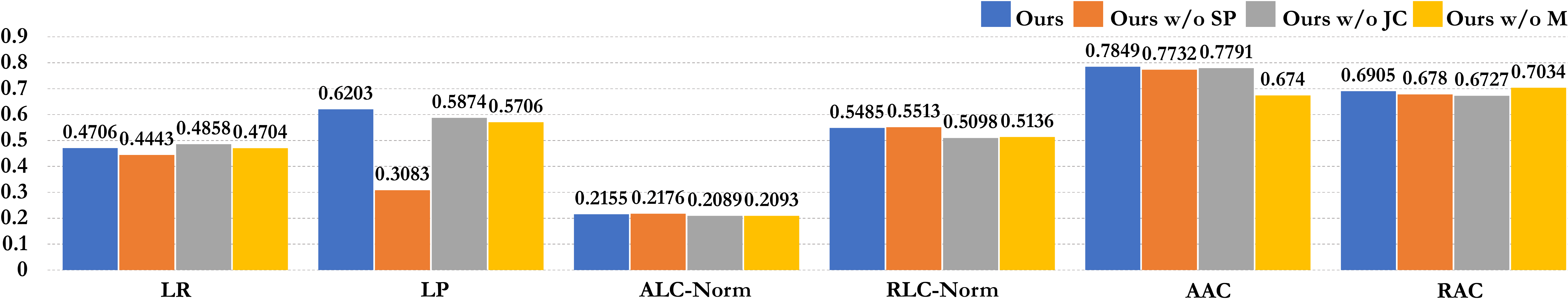}
\caption{Performance of ablation study of text-to-layout generation in \textbf{LR}, \textbf{LP}, \textbf{ALC-Norm}, \textbf{RLC-Norm}, \textbf{AAC}, and \textbf{RAC} on COCO-stuff dataset. ALC-Norm and RLC-Norm are the values of ALC and RLC parameterized by Gaussian kernel function.}
\label{fig:ablation}
\end{figure*}

\ljdcomment{
When calculating LC, the generated coordinates and the annotated ones of the o-th object should be matched. 
However, the objects are generated in random order at the first stage.
Hence, we first match the bounding boxes of the same category. 
When the same category has several objects, we select the group with the lowest ALC from all combinations.
}

\noindent\textbf{Area Consistency (AC)}, which measures the consistency between the area of the predicted object bounding boxes and the area of groundtruth object bounding boxes. 
Similar to the definition of \textbf{LC}, \textbf{AC} is also defined considering two types of measurements: 1) Absolute Area Consistency (AAC) measuring the difference between area of predicted objects and the corresponding groundtruth objects; 2) Relative Area Consistency (RAC) which measures the consistency of pairwise size relationship. Suppose the bounding box area for the $i$-th predicted object and the corresponding groundtruth object in $\mathbf{m} = \mathbf{d} \cap \hat{\mathbf{d}}$ are $\hat{u}_i$ and $u_i$, then AAC and RAC are defined as:
\begin{equation}
\resizebox{0.8\hsize}{!}{$
\begin{split}
& \text{AAC} = 1 - \frac{1}{\mid \mathbf{m} \mid}\sum_{i \in \mathbf{m}}\frac{|\hat{u}_i - u_i|}{u_s},\\
&\text{RAC} = \frac{\sum_{i \in \mathbf{m}}\sum_{j\in \mathbf{m} \ \& \ i \neq j}
\big(1- | \mathbbm{1}_{u_i > u_j} - \mathbbm{1}_{\hat{u}_i > \hat{u}_j} | \big)
}{\mid \mathbf{m} \mid\times(\mid \mathbf{m} \mid-1)},
\end{split}
$}
\label{eqn:RAA}
\end{equation}

where $\mathbbm{1}$ is the indicator function. Finally, we define Area Consistency (\textbf{AC}) by combining both AAC and RAC  based on a Gaussian kernel function (with parameter $\sigma_3$):
\begin{equation}
\text{AC} = \gamma_{ac}\text{AAC} + (1 - \gamma_{ac})\text{RAC}.
\label{eqn:ADS}
\end{equation}

\begin{figure}[t]
\centering
\includegraphics[height=4.7cm]{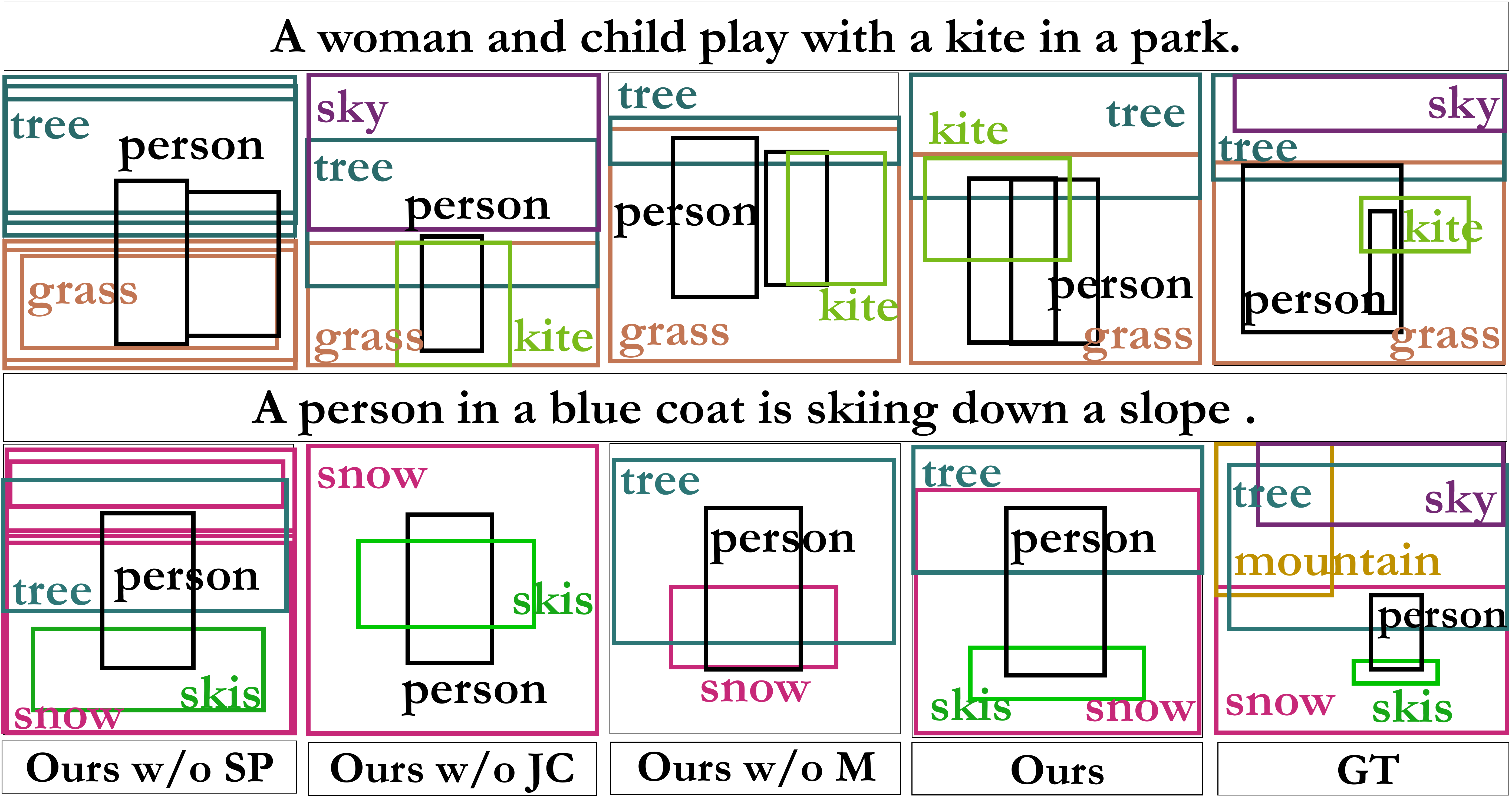}
\caption{Qualitative comparison of generated Layouts between different variants of our model for ablation study on COCO-stuff dataset.}
\label{fig:layput_vis_ablation}
\end{figure}
\textbf{LR} and \textbf{LP} are defined to measure the correctness of predicted object categories while \textbf{LC} and \textbf{AC} measures the quality of the bounding boxes including location and shape. Taking into account all four measurements, we calculate the Layout Quality Score ($\textbf{LQS} \in [0, 4]$) by:
\begin{equation}
\resizebox{0.5\hsize}{!}{$
    \text{LQS} = \text{LR} + \text{LP} + \text{LC} + \text{AC}.
    $}
\label{eqn:LAS}
\end{equation}

Although both LC and AC consist of absolute and relative parts. 
Since a text input can be consistent with multiple different layouts, relative consistency is more important than absolute consistency. Therefore, in this paper, we set both $\gamma_{\text{lc}}$ and $\gamma_{\text{ac}}$ to 0.25. 
Subsequently, with 256x256 image area resolution, $u_s$ is set to 80 to obtain smoothing values of LC. By adjusting these hyperparameters, LQS can perform quantitative evaluations at different image resolutions and different types of layout generation tasks.

Besides, we also employ LostGAN\_V2~\cite{sun2020learning}, a state-of-the-art layout-to-image synthesizer, to synthesize images from a generated layout and calculate \textbf{IS}, \textbf{FID}, Diversity Score (\textbf{DS}) and Classification Accuracy Score (\textbf{CAS}) of the images as four indirect  metrics for evaluating the layout.
\textbf{DS} calculates perceptual similarity between two image datasets in deep feature space. In the paper, LPIPS metric~\cite{zhang2018unreasonable} is adopted to compute perceptual similarity between two images. \textbf{CAS} is used to evaluate the recognizability of objects in the generated images. To be specific, we adopt a Resnet-50 model trained on object crops obtained from the corresponding train dataset to classify object crops from the synthesized image.

\subsection{Evaluation of Generated Layouts.}
\begin{figure*}[t]
\centering
\includegraphics[height=16.2cm]{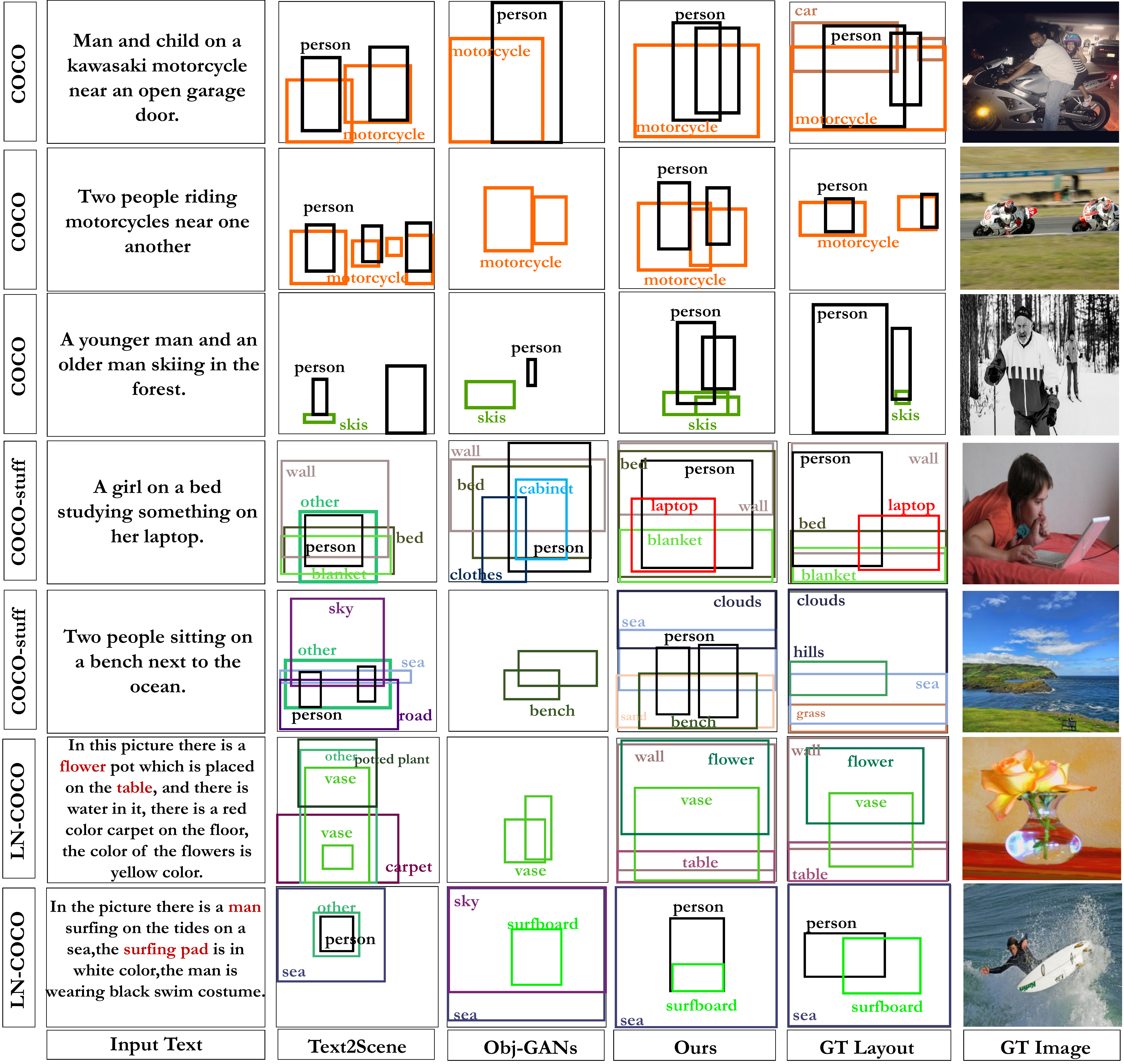}
\caption{Qualitative comparison of generated Layouts between our method and other two methods on three different datasets.}
\label{fig:layput_vis_previous}
\end{figure*}

\begin{table*}[h]
\begin{center}
\caption{Performance of different methods for text-to-layout generation on three different datasets.}
\label{table:bbox_evaluation}
\resizebox{1.0\linewidth}{!}{
\begin{tabular}{l|l|c|c|ccc|ccc|c|cccc}
\toprule
Dataset & Model & LR $\uparrow$ & LP $\uparrow$ & ALC $\downarrow$ & RLC $\downarrow$ & LC $\uparrow$ & AAC $\uparrow$ & RAC $\uparrow$ & AC $\uparrow$ & LQS $\uparrow$ & IS $\uparrow$ & FID $\downarrow$ & DS $\downarrow$ & CAS $\uparrow$\\
\midrule
\multirow{3}*{\tabincell{c}{COCO}}
& Obj-GANs & 46.58\% & 68.37\% & 63.4494 & 97.9462 & 0.5370 & 0.8268 & 0.6859 & 0.7211  & 2.4076  & - & - & - & -  \\
~ & Text2Scene & \textbf{49.16}\% & 61.24\% & 91.3040 & 102.4082 & 0.4609 & \textbf{0.8634} & 0.6809 & 0.7265  & 2.2914  & - & - & - & -  \\
~ & Ours & 47.44\% & \textbf{75.75}\% & \textbf{57.2088} & \textbf{88.5594} & \textbf{0.6000} & 0.8481 & \textbf{0.6964} & \textbf{0.7343}  & \textbf{2.5662}  & - & - & - & -  \\
\midrule
\multirow{3}*{COCO-stuff}
& Obj-GANs & 33.36\% & 48.09\% & 54.4306 & 77.3151 & 0.6685 & \textbf{0.7915} & 0.6144 & 0.6587  & 2.1417  & 10.75 & 88.28 & 0.7288 & 0.3765  \\
~ & Text2Scene & 44.30\% & 48.33\% & 64.0896 & 65.1006 & 0.7200 & 0.7719 & 0.6811 & 0.7038  & 2.3501 & 5.79 & 126.08 & 0.7149 & 0.2410  \\
~ & Ours & \textbf{47.06}\% & \textbf{62.03}\% & \textbf{43.6186} & \textbf{63.2825} & \textbf{0.7640} & 0.7849 & \textbf{0.6905} & \textbf{0.7141}  & \textbf{2.5690} &  \textbf{19.03} & \textbf{39.40} & \textbf{0.6944} & \textbf{0.6228} \\
\midrule
\multirow{3}*{LN-COCO}
& Obj-GANs & 27.65\% & 41.11\% & 54.5414 & 81.7542 & 0.6431 & \textbf{0.7899} & 0.5796 & 0.6322  & 1.9629 & 9.65 & 128.06 & 0.7165 & 0.4018  \\
~ & Text2Scene & \textbf{49.26}\% & 53.14\% & 66.6123 & 68.2388 & 0.6981 & 0.7786 & 0.6869 & 0.7098  & 2.4319  & 5.69 & 100.63 & 0.7110 & 0.2664  \\
~ & Ours & 46.61\% & \textbf{62.70}\% & \textbf{41.6681} & \textbf{59.6552} & \textbf{0.7862} & 0.7795 & \textbf{0.6893} & \textbf{0.7118}  & \textbf{2.5912} & \textbf{19.27} & \textbf{40.04} & \textbf{0.6894} & \textbf{0.6719}  \\
\bottomrule
\end{tabular}}
\end{center}
\end{table*}

 In this section we evaluate the proposed LayoutTransformer for text-to-layout Generation both quantitatively and qualitatively. 
 we first conduct ablation study to evaluate the effectiveness of each module in LayoutTransformer. Next, we compare our LayoutTransformer with other methods for text-to-layout generation. Finally, we visualize the synthesized layouts for several examples to gain more insight into effectiveness of LayoutTransformer.
 
\subsubsection{Ablation Study} \quad
\label{Sec:ablation}

\noindent{\textbf{Effect of sequential object prediction.}
To investigate the effect of sequential object prediction mechanism, we evaluate the variant of LayoutTransformer(Ours), termed as Ours w/o SP, which predicts the objects in parallel. Thus the dependencies between objects are not modeled. 
As a result, layouts generated by Ours w/o SP have redundant bounding boxes('tree', 'grass' and 'snow') as shown in Figure~\ref{fig:layput_vis_ablation}.
The quantitative results in Figure~\ref{fig:ablation} also show that Ours w/o SP performs much worse than LayoutTransformer (especially on LP).
The sequential object prediction enables the LayoutTransformer to perceive the previously generated objects and effectively reduces the repeated generation of objects of the same category based on the semantics of the input text.}

\noindent{\textbf{Effect of joint classification of object category and position.}
Inspired by the strong correlation between object category and position, joint classification can perform better location estimation of object.
To investigate the effect of such design, we evaluate the performance of the variant of our model that predicts the object category and position (in grid) separately, denoted as Ours w/o JC. Figure~\ref{fig:ablation} shows that the performance of Ours w/o JC is distinctly lower than original LayoutTransformer (especially on LP and RLC).
Compared to the full LayoutTransformer, interactions between objects ('person/kite' and 'person/skis') are incorrect as shown in Figure~\ref{fig:layput_vis_ablation}, which validates that joint prediction can achieve more accurate estimation of spatial relationships between objects.}

\noindent\textbf{Incorporating multi-caption input vs single-caption input.} To investigate the effect of incorporating multi-caption input, we evaluate the variant of LayoutTransformer, termed as Ours w/o M, which only uses single caption as input in the training stage. 
Due to the diversity of textual expressions, the text encoder of Ours w/o M cannot precisely extract the semantic information in the input text. Figure~\ref{fig:ablation} shows that Ours w/o M also performs worse than the original version on LP, RLC and AAC. Besides, Figure~\ref{fig:layput_vis_ablation} also manifests that the generated layouts by Ours w/o M cannot learn reasonable layout between objects.

\subsubsection{Comparison with State-of-the-art Methods} \quad

\noindent{\textbf{Quantitative Evaluation.}
Table~\ref{table:bbox_evaluation} presents the experimental results of our LayoutTransformer and other state-of-the-art methods for text-to-layout generation: Text2Scene~\cite{tan2019text2scene} and Obj-GANs~\cite{li2019object} on three datasets. 
In addition to LR and AAC, our LayoutTransformer achieves the best performance in other metrics and outperforms other methods significantly in terms of LP, LC, IS, and FID, which is owing to joint contributions from all three modules we proposed.
Particularly, the enhancement of LC shows that our method can generate layouts with more correct spatial relationships.
The performance of IS, FID, DS, and CAS also prove that the layout generated by our method can be used to synthesize high quality images containing more recognizable objects.
Text2Scene could achieve the best LR on COCO and LN-COCO datasets due to the generation of redundant bounding boxes as shown in Figure~\ref{fig:layput_vis_previous}.}

\noindent\textbf{Qualitative Evaluation.}
The qualitative comparisons in Figure~\ref{fig:layput_vis_previous} reveal that our method is able to generate more reasonable layouts between objects than other methods on three different datasets. Other methods either fail to predict correct object categories, or cannot correctly establish spatial relationships between different objects. 

In the COCO dataset, we observe that Obj-GANs and Text2scene either fail to predict correct object categories, or cannot correctly establish spatial relationships between different objects. The results in the first and second rows of Figure~\ref{fig:layput_vis_previous} show that our model can synthesize layouts that reflect subtle semantic differences of input captions (riding the same motorcycle vs two different motorcycles).
The results in the third row of Figure~\ref{fig:layput_vis_previous} also demonstrate that our method is not simply overfitting the dataset but can comprehensively understand the input caption and generate a reasonable layout.
The COCO-stuff dataset contains more categories which increases the challenge of text-to-layout synthesis but our model still can generate reasonable layouts.
In the LN-COCO dataset, the longer captions further increase the challenge of semantic parsing of text encoder.
Obj-GANs and Text2Scene lack the ability to parse semantics of long input captions to correctly predict object categories as illustrated in Figure~\ref{fig:layput_vis_previous}.
Since the outstanding long-range semantic parsing capability of our text encoder, our model is still able to generate more reasonable layouts.

\begin{table*}[t]
\begin{center}
\caption{Quantitative results of different text-to-image synthesis models on COCO dataset.}
\label{table:image_evaluation}
\resizebox{0.9\linewidth}{!}{
\begin{tabular}{l|ccccccc|c}
\toprule
Model & IS $\uparrow$ & IS w/o DAMSM $\uparrow$ & FID $\downarrow$ & R-precision $\uparrow$ & ClipScore $\uparrow$ & SOA-C $\uparrow$ & SOA-I $\uparrow$ & Rank-1 ratio $\uparrow$ \\
\midrule
AttnGAN~\cite{xu2018attngan} & 23.61 & 12.22 & 33.10 & 53.36 & 68.40 & 25.88 & 39.01 & 3.36\%  \\
DM-GAN~\cite{zhu2019dm} & 32.32 & 11.65 & 27.34 & 69.10 & 72.30 & 33.44 & 48.03 & 3.38\%  \\
DF-GAN~\cite{tao2020df} & - & 17.25 & 19.32 & 39.06 & 65.34 & 17.77 & 30.42 & \underline{10.88\%} \\
DAE-GAN~\cite{ruan2021dae} & \underline{33.11} & 10.57 & 56.85 & 66.92 & 71.31 & 30.45 & 45.08 & 7.12\% \\
TIME~\cite{liu2020time} & 30.85 & - & 31.14 & - & - & 32.78 & - & - \\
Huang~\cite{huang2021unifying} & - & \textbf{25.31} & 29.90 & \underline{69.90} & \underline{77.20} & \underline{38.12} & \underline{53.31} & - \\
VQ-Diffusion-B~\cite{gu2022vector} & - & - & 19.75 & - & - & - & - & - \\
CSM-GAN~\cite{CSM-GAN} & 26.77 & - & 33.48 & - & - & - & - & - \\
CGL-GAN~\cite{CGL-GAN} & 13.62 & - & 37.12 & - & - & - & - & - \\
KD-GAN~\cite{KD-GAN} & 34.01 & - & 23.92 & - & - & - & - & - \\
DM-GAN-MDD~\cite{MDD} & 34.46 & - & 24.30 & - & - & - & - & - \\
\midrule
Obj-GANs\cite{li2019object} & 24.09 & 11.34 & 36.52 & 60.05 & 70.26 & 27.14 & 41.24 & 5.64\% \\
OP-GAN~\cite{hinz2019semantic} & 27.88 & 10.32 & \underline{24.07} & 63.81 & 71.14 & 35.85 & 50.47 & 6.76\% \\
R-GAN~\cite{qiao2021r} & - & - & 24.60 & - & - & - & - & - \\
\midrule
Ours & \textbf{38.43} & \underline{22.95} & \textbf{15.85} & \textbf{78.40} & \textbf{78.82} & \textbf{38.59} & \textbf{57.05} & \textbf{54.44}\%\\
\bottomrule
\end{tabular}}
\end{center}
\end{table*}

\begin{figure}[t]
\centering
\includegraphics[height=5cm]{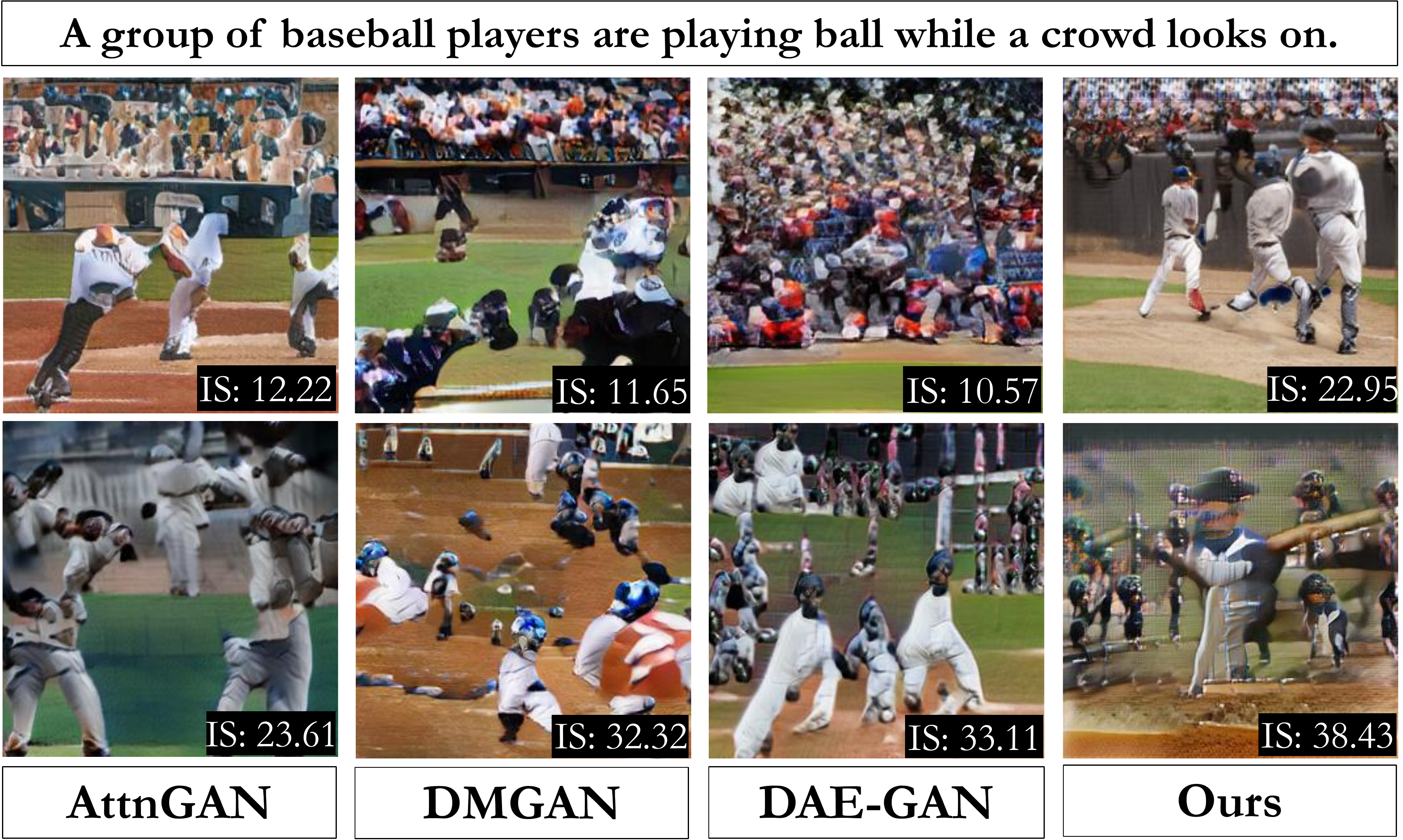}
\caption{Visualization of synthesized images by different models with (\textbf{bottom} row) and without DAMSM (\textbf{top} row) respectively.}
\label{fig:compare_damsm}
\end{figure} 

\subsection{Evaluation of Synthesized Images}
Next we compare our method with other state-of-the-art methods for text-to-image synthesis.

\noindent\textbf{Quantitative Evaluation.}
Following previous work, we report validation results by generating images for 30,000 random captions. Table~\ref{table:image_evaluation} reports the experimental results in terms of IS, FID, R-precision, ClipScore, SOA-C, and SOA-I of different methods for text-to-image synthesis on the COCO dataset. 
Here we present two versions of performance in IS: 1) using DAMSM~\cite{xu2018attngan} (IS), which is a loss function for pushing for the text-to-image alignment and 2) performance without using DAMSM (IS w/o DAMSM). We report these two versions of performance since DAMSM has a substantial impact and bias towards IS score due to the same pre-trained Inception-V3~\cite{szegedy2016rethinking} module for image encoding employed by both DAMSM and IS calculation. 
To investigate whether DAMSM can improve the quality of the synthesized images for a same model, we visualize the synthesized images by different models, with and without DAMSM respectively in Figure~\ref{fig:compare_damsm}. We observe that DAMSM degenerates the performance for almost all models, probably because DAMSM leads to the overfitting on IS score.
The results in Table~\ref{table:image_evaluation} show that our model achieves the second best performance on IS w/o DAMSM and best performance on other metrics and outperforms other methods significantly in terms of FID and R-precision which is owing to the guidance of layout in text-to-image synthesis.

\noindent\textbf{Human Evaluation.}
As a complement to the standard evaluation metrics, we also perform human evaluation to compare our method with other state-of-the-art methods. Specifically, we randomly select 50 samples from the test set and ask human subjects to compare the quality of the synthesized images by different methods and vote for the best one, considering both the text-image consistency and the quality of object layout.
We calculate the rank-1 ratio for each method as the metric as shown in Table~\ref{table:image_evaluation}. Our method achieves $54.44\%$ votes from collected 50 human subjects, which outperforms other methods significantly.

\begin{figure}[t]
\centering
\includegraphics[height=4.8cm]{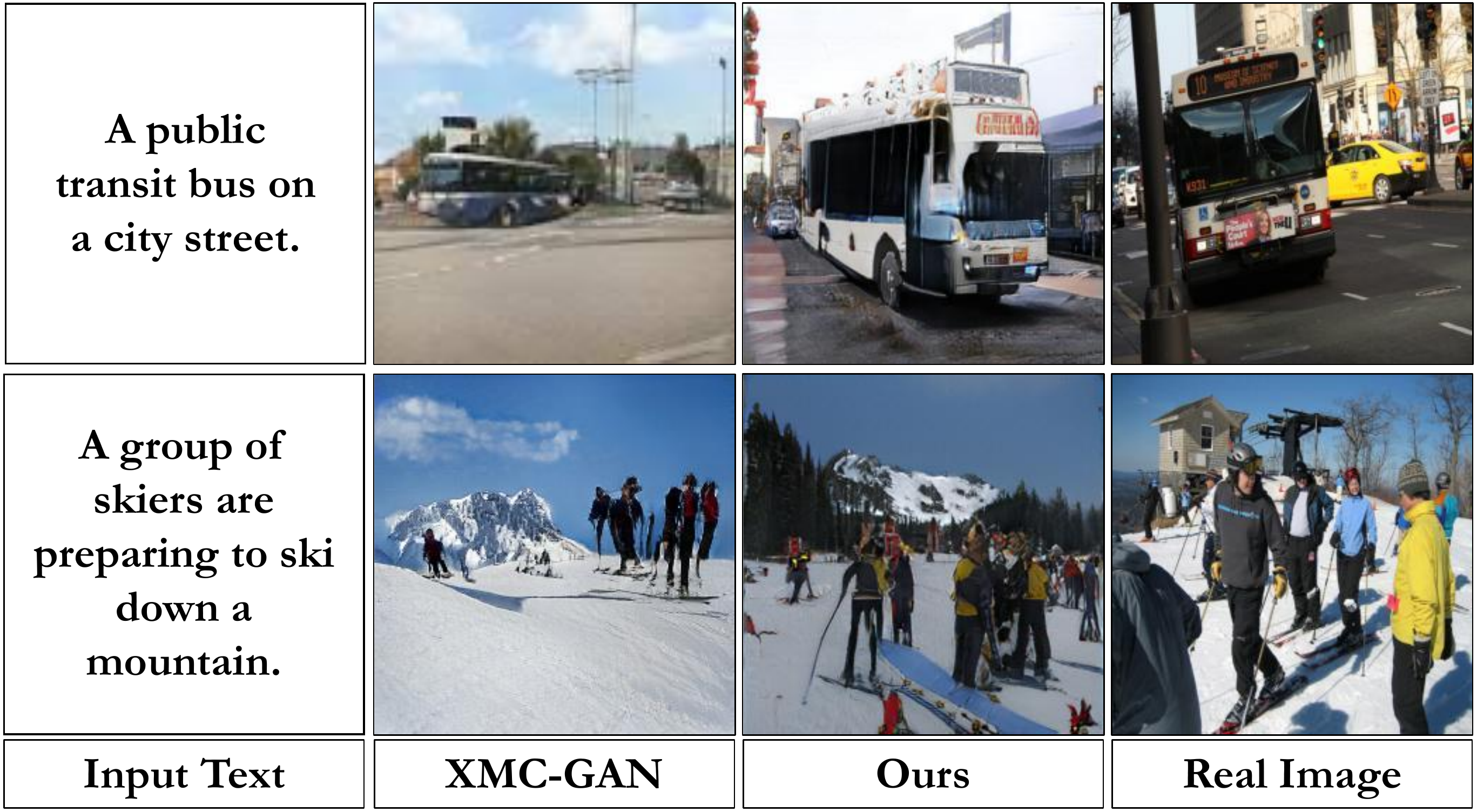}
\caption{Qualitative comparison between our model with XMC-GAN. The XMC-GAN results are extracted from their paper.}
\label{fig:compare_xmc}
\end{figure}

\begin{figure}[t]
\begin{center}
\includegraphics[width=1.0\linewidth]{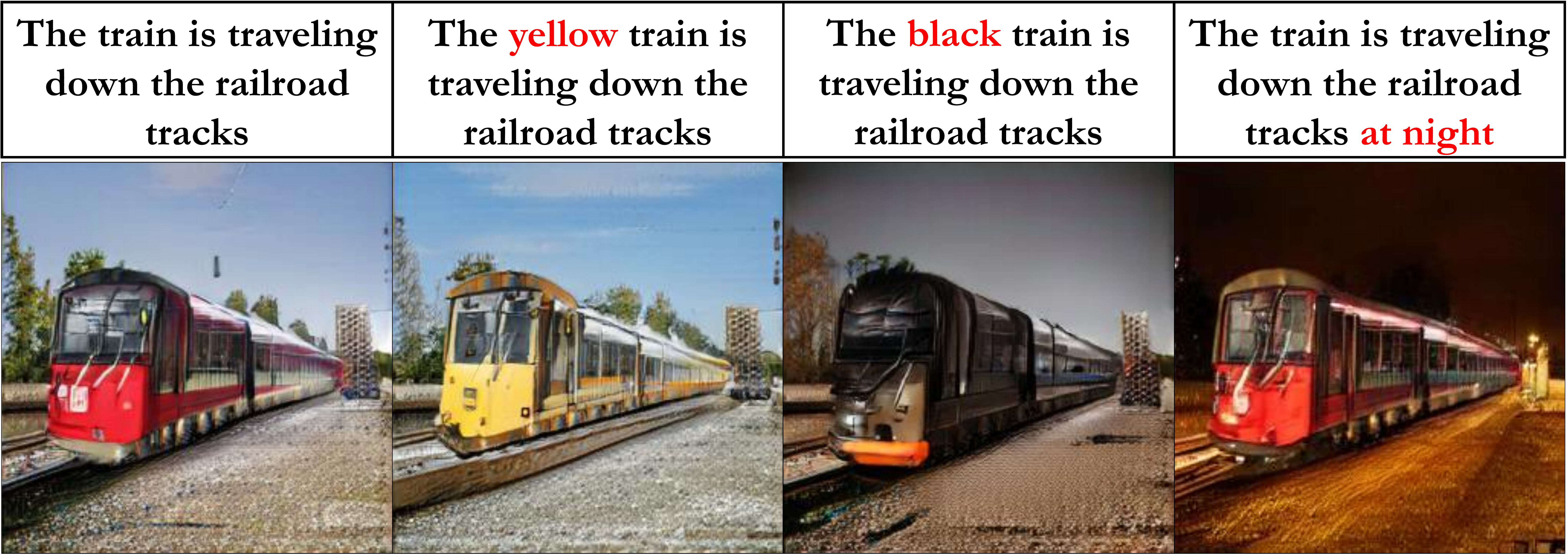}
\end{center}
\caption{Synthesized images by our model given four close-semantic descriptions. }
\label{fig:TALIS}
\end{figure}

\begin{figure*}[t]
\centering
\includegraphics[height=14cm]{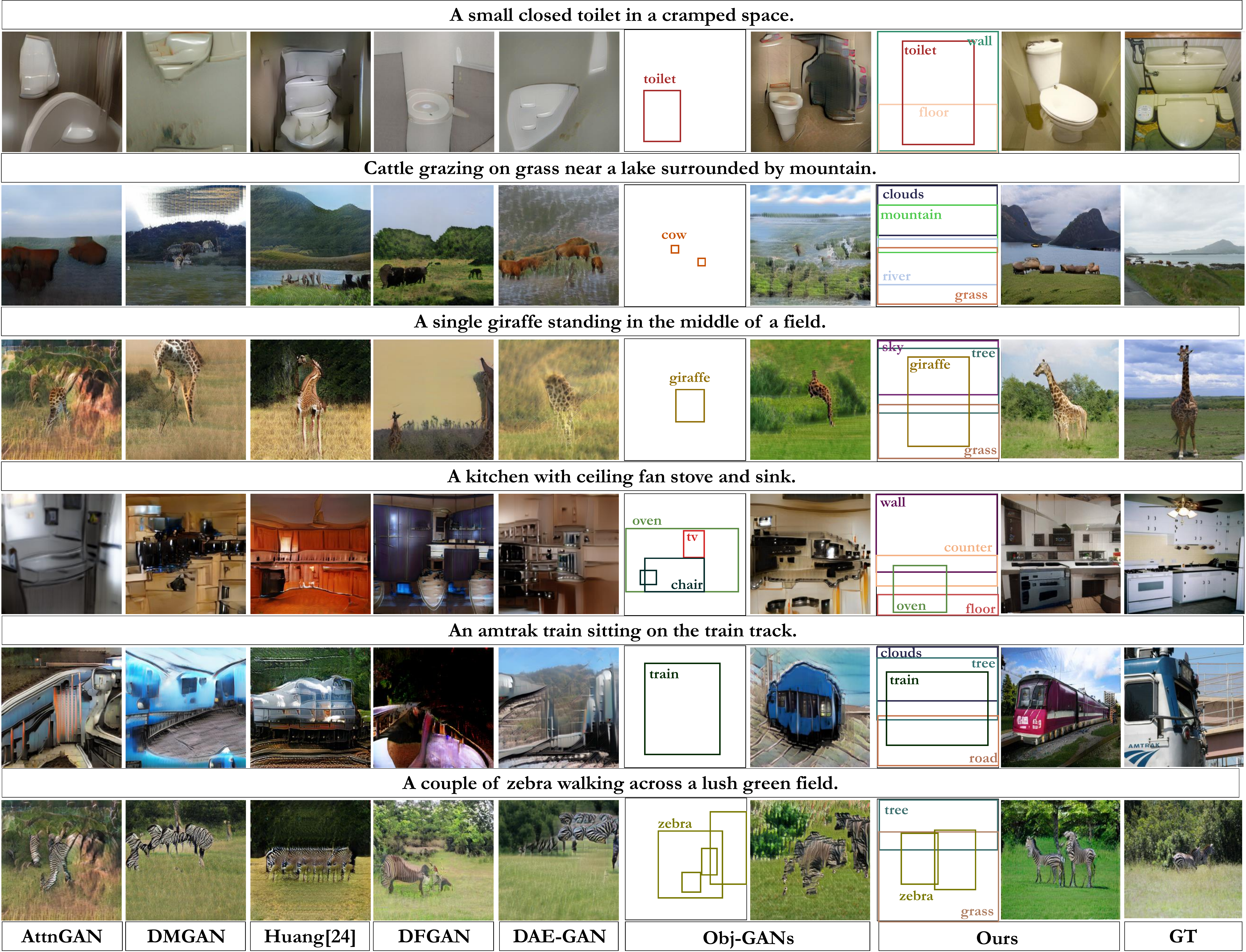}
\caption{Qualitative comparison between our model with other state-of-the-art models for text-to-image synthesis.}
\label{fig:compare_image_vis}
\end{figure*}

\noindent\textbf{Qualitative Evaluation.}
To have a qualitative comparison, we visualize the synthesized images by our models and other state-of-the-art models in Figure~\ref{fig:compare_image_vis}. Note that Obj-GANs and our method synthesize images via generating intermediate layout while other methods perform text-to-image synthesis directly. The visualized results clearly show that the synthesized images by our method are much better than the results of other methods, in terms of both image quality and the reasonability of object layouts.
Specifically, the object locations in the second (lake, mountain, and cattle) and fourth row (stove and sink) of the Figure~\ref{fig:compare_image_vis} show that the images synthesized by our method have a more reasonable layout.
Besides, the objects in the images generated by our method also have a higher authenticity than other models (toilet, giraffe, and train).

We also compare our results with XMC-GAN~\cite{zhang2021cross} in Figure~\ref{fig:compare_xmc}.
XMC-GAN effectively improves the quality of the synthesized image by using cross-modal contrastive Learning strategy on the task of text-to-image generation.
Since it does not release the pre-trained model and is difficult to reproduce results in their paper.
Therefore, we only perform a qualitative evaluation of XMC-GAN.
Compared to XMC-GAN which requires training 1000 epochs on a 32-core Pod slice of Google Cloud TPU v3 devices, our method only requires training 150 epochs on 4 Tesla V100 GPUs.
However, our approach also has achieved competitive performance as shown in Figure~\ref{fig:compare_xmc}.

\subsection{Investigation on Effectiveness of TALIS}
Compared to LostGAN\_{V2}, our \emph{TALIS} learns the textual-visual semantic alignment between the input text and the synthesized image by the proposed Object-Oriented Text Encoder. 
Table~\ref{table:TALIS} reports the results of the quantitative comparison. 
Our method outperforms LostGAN\_V2 in all four metrics and advances the FID significantly (from 39.40 to 35.05).
Besides, we also visualize an example in Figure~\ref{fig:TALIS}, in which the semantics of the four textual descriptions are similar but not exactly the same. Our model is able to synthesize images that reflect these subtle semantic differences.

\begin{table}[h]
\begin{center}
\caption{Quantitative comparison results for LostGAN\_V2 on COCO-stuff dataset. We feed the generated layout by our LayoutTransformer
into LostGAN\_V2 as input}
\label{table:TALIS}
\resizebox{0.8\linewidth}{!}{
\begin{tabular}{l|cccc}
\toprule
Model & IS $\uparrow$  & FID $\downarrow$ & DS $\downarrow$ & CAS $\uparrow$ \\
\midrule
LostGAN\_V2$^*$~\cite{sun2020learning} & 19.03 & 39.40 & 0.6944 & 0.6228  \\
\midrule
Ours & \textbf{20.35} & \textbf{35.05} & \textbf{0.6920} & \textbf{0.6314} \\
\bottomrule
\end{tabular}}
\end{center}
\end{table}

\subsection{Discussion of Limitations}
\label{sec:Limitations}
While our method can generate plausible layouts in realistic scenarios given textual descriptions, the layout-to-image synthesis for our method remains a challenging problem. This particularly results from quite diverse appearances for some categories of objects in COCO dataset. We will continue to investigate this problem in the future work.

\section{Conclusion}
In this paper we perform text-to-image synthesis by generating an intermediate layout to bridge the input text and the synthesized image. We have presented LayoutTransformer for text-to-layout generation, which formulates the layout generation as a task of sequential object prediction. Thus the relationships between objects can be learned by modeling the sequential dependencies between object predictions. The generated layout is further leveraged to guide image synthesis performed by the proposed layout-to-image synthesizer, which focuses on learning textual-visual semantic alignment per object in the layout. Extensive experiments validate the effectiveness of the proposed method.

\bibliographystyle{IEEEtran}
\bibliography{egbib}

\newpage

\vfill

\end{document}